\documentclass{article}

\usepackage[final]{neurips_2024}
\usepackage[utf8]{inputenc} 
\usepackage[T1]{fontenc}    
\usepackage{hyperref}       
\usepackage{url}            
\usepackage{booktabs}       
\usepackage{amsfonts}       
\usepackage{nicefrac}       
\usepackage{microtype}      
\usepackage{xcolor}         
\usepackage{colortbl}
\usepackage{makecell}
\usepackage{graphicx}
\usepackage{wrapfig}
\usepackage{amsmath}
\usepackage{amssymb}
\usepackage{multirow}
\usepackage{algorithm}
\usepackage{algpseudocode}
\usepackage{tikz}
\usepackage{xspace}
\usepackage{amsfonts}
\usepackage{array,multirow}
\usepackage{pgfplots}
\usepackage{verbatim}
\usepackage{caption}
\usepackage{multirow}
\usepackage{bm}
\usepackage{multirow}
\usepackage{booktabs}
\usepackage{color}
\usepackage{framed}
\usepackage{stfloats}
\usepackage{iitem}
\usepackage{amssymb}
\usepackage{pifont}
\usepackage{arydshln}
\usepackage{enumitem}
\usepackage{wrapfig}
\usepackage{algorithm}
\usepackage{algpseudocode}
\usepackage{arydshln}
\usepackage{multirow}
\usepackage{threeparttable}
\usepackage{bbding}
\usepackage{pifont}
\usepackage{wasysym}
\usepackage{amssymb}

\usepackage[utf8]{inputenc} 
\usepackage[T1]{fontenc}    
\usepackage{hyperref}       
\usepackage{url}            
\usepackage{booktabs}       
\usepackage{amsfonts}       
\usepackage{nicefrac}       
\usepackage{microtype}      
\usepackage{xcolor}         

\title{SelectIT: Selective Instruction Tuning for LLMs \\ via Uncertainty-Aware Self-Reflection}

\author{Liangxin Liu$^1$~~~
        Xuebo Liu$^1$\thanks{~~Corresponding Author}~~~
        Derek F. Wong$^2$~~~
        Dongfang Li$^1$~~~\\
        \textbf{Ziyi Wang$^1$~~~
        Baotian Hu$^1$~~~
        Min Zhang$^1$}\\
    $^1$Institute of Computing and Intelligence, Harbin Institute of Technology, Shenzhen, China \\
    $^2$NLP$^2$CT Lab, Department of Computer and Information Science, University of Macau \\
\texttt{lliangxin967@gmail.com, \{liuxuebo,hubaotian,zhangmin2021\}@hit.edu.cn} \\
   \texttt{derekfw@um.edu.mo, \{crazyofapple,ziyiwang676\}@gmail.com}  }

\begin{document}

\maketitle

\begin{abstract}
Instruction tuning (IT) is crucial to tailoring large language models (LLMs) towards human-centric interactions.
Recent advancements have shown that the careful selection of a small, high-quality subset of IT data can significantly enhance the performance of LLMs. 
Despite this, common approaches often rely on additional models or data, which increases costs and limits widespread adoption.
In this work, we propose a novel approach, termed \textit{SelectIT}, that capitalizes on the foundational capabilities of the LLM itself. 
Specifically, we exploit the intrinsic uncertainty present in LLMs to more effectively select high-quality IT data, without the need for extra resources. 
Furthermore, we introduce a curated IT dataset, the \textit{Selective Alpaca}, created by applying SelectIT to the Alpaca-GPT4 dataset.
Empirical results demonstrate that IT using Selective Alpaca leads to substantial model ability enhancement.
The robustness of SelectIT has also been corroborated in various foundation models and domain-specific tasks. 
Our findings suggest that longer and more computationally intensive IT data may serve as superior sources of IT, offering valuable insights for future research in this area. 
Data, code, and scripts are freely available at \url{https://github.com/Blue-Raincoat/SelectIT}.
\end{abstract}

\section{Introduction}

\begin{wrapfigure}{r}{6.5cm}
\vspace{-0.4cm}
    \centering
    \includegraphics[scale=1]
{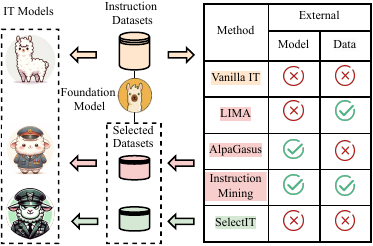}
\caption{Existing advanced data selection strategies rely heavily on external models or data; however, SelectIT effectively overcomes this limitation.}
    \label{fig:model_competence}
\vspace{-0.4cm}

\end{wrapfigure}

Large language models (LLMs) have attracted much attention due to their impressive capabilities in following instructions and solving intricate problems~\citep{touvron2023LLaMA,touvron2023LLaMA1,achiam2023gpt,penedo2023refinedweb}.
A crucial aspect of enhancing LLMs' performance is instruction tuning (IT), which involves the supervised adjustment of LLMs using pairs of instructional data, essential for refining the models' ability to accurately respond to human instructions. 
Recent groundbreaking research, such as the LIMA~\citep{zhou2023lima}, highlights the critical importance of instructional data quality over quantity. 
Contrary to the approach of merely increasing the dataset size, a carefully selected, smaller dataset of higher quality can significantly improve LLMs' performance.

Despite the development of various high-quality data selection methods, they often depend on external resources, limiting wider implementation.
\textit{External Model}:~\citet{chen2023alpagasus,liu2023makes} propose the employment of closed-source LLMs to evaluate or rank IT data.
To circumvent the closed-source limitations,~\citet{li2023quantity, li2023self, kung2023active} recommend fine-tuning open-source LLMs, which requires more computational resources.
\textit{External Data}:~\citet{cao2023instruction} split all mixed data into several bins and fully trained the models to evaluate different indicators of high-quality IT data.
Despite these advancements, the challenge of precise and efficient high-quality data selection without external resources remains unresolved.

In this paper, we introduce \textit{SelectIT}, a novel approach designed to enhance IT data selection by fully leveraging the foundation model itself, eliminating the need for external resources.
SelectIT employs different grain uncertainty of LLMs: token, sentence, and model, which can effectually improve the accuracy of IT data selection.
We first use the foundation model itself to rate the IT data from 1 to $K$ based on the uncertainty of various tokens.
Next, we use sentence-level uncertainty to improve the rating process by exploiting the effect of different prompts on LLMs.
At a higher model level, we utilize the uncertainty between different LLMs, enabling a collaborative decision-making process for IT data selection.
By applying SelectIT to the original Alpaca, we curate a compact and superior IT dataset, termed \textit{Selective Alpaca}.

Experimental results show that SelectIT outperforms existing high-quality data selection methods, improving LLM's performance on the open-instruct benchmark~\citep{wang2023far}.
Further analysis reveals that SelectIT can effectively discard abnormal data and tends to select longer and more computationally intensive IT data.
The primary contributions of SelectIT are as follows:
\begin{itemize}

\item We propose SelectIT, a novel IT data selection method which exploits the uncertainty of LLMs without using additional resources.

\item We introduce a curated IT dataset, Selective Alpaca, by selecting the high-quality IT data from the Alpaca-GPT4 dataset.

\item SelectIT can substantially improve the performance of LLMs across a variety of foundation models and domain-specific tasks.

\item Our analysis suggests that longer and more computationally intensive IT data may be more effective, offering a new perspective on the characteristics of optimal IT data.

\end{itemize}

\section{Related Work}
\paragraph{Instruction Tuning Dataset}
Recent empirical research highlights the substantial benefits of fine-tuning LLMs on specialized datasets containing instructions and responses, significantly enhancing their generalization capabilities and responsiveness to new questions~\citep{chung2022scaling,longpre2023flan,honovich2022unnatural,sun2023principle}.
FLAN~\citep{wei2021finetuned} reformulates traditional natural language processing tasks as instructions formats, thereby improving model performance.
Alpaca~\citep{Alpaca,peng2023instruction} exemplifies the effectiveness of merging a select set of manual instruction seeds with advanced LLMs, like text-davinci-003 or GPT-4, to compile a comprehensive dataset.
Similarly, Vicuna~\citep{chiang2023vicuna} leverages 70,000 conversations from ChatGPT interactions, benefiting from the diverse data types and structures within these dialogues.
WizardLM~\citep{xu2023wizardlm} introduces a novel approach by using LLMs to automatically generate open-domain instructions of varying complexities, achieving controlled instructional difficulty variation.
However, LIMA~\citep{zhou2023lima} demonstrates that only 1$K$ high-quality IT data can match or exceed the performance of LLMs fine-tuned on larger IT datasets, presenting a promising direction for future research.

\paragraph{Instruction Data Selection}
The recognition of IT data quality's superiority over quantity in the context of IT is well-established, yet the efficient and precise identification of high-quality data continues to be a challenging frontier for research. 
One straightforward approach is utilizing the closed-source advanced LLMs for IT data evaluation and selection~\citep{chen2023alpagasus,liu2023makes}. 
To circumvent the constraints associated with closed-source, existing research opt to fine-tune LLMs directly to select high-quality IT data~\citep{li2023self,kung2023active}.
\citet{li2023one,gururangan2020don,chen2024skill,cao2023instruction} use pre-defined notions of useful data or other IT datasets to develop a data quality assessment framework.
\citet{li2023quantity} propose training a specialized model and utilizing two unique, condition-based losses on this for a comprehensive IT data selection.
\citet{wu2023self} explore where data selection is informed by the similarity of samples within the embedding space of a fine-tuned model.
N-gram features~\citep{xie2023data} or model gradients \citep{xia2024less,han2023understanding}  are also important features for selecting high-quality data in fine-tuned LLMs.
However, the methods described above depend, to varying degrees, on supplementary datasets, the use of closed-source models, or open-source models that have been specially fine-tuned, which results in increased consumption of resources and potentially limits the broader impact.

 \begin{figure*}[t]
	\centering
 \scalebox{1.1}{
	\includegraphics[]{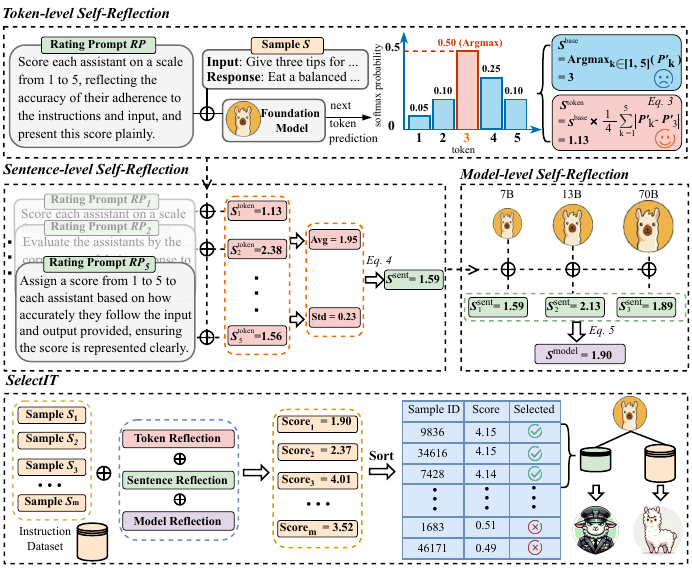}}
	\caption{{Overall framework of SelectIT. In Token-level Self-Reflection, we employ the foundation model to rate the IT data from 1 to $K$. In Sentence-level Self-Reflection, we leverage the uncertainty of varied prompts on LLMs to enhance the rating process. In Model-level Self-Reflection, we harness uncertainty among different LLMs to facilitate a collaborative decision-making process in selecting IT data. Finally, different levels of self-reflection are reasonably combined into SelectIT, which can effectively select high-quality IT data without relying on additional resources.}}
	\label{table:main_pic}
\end{figure*}

\section{Our SelectIT Method}
Utilizing advanced LLMs for the sample evaluation is a widely adopted approach in the IT data selection~\citep{chen2023alpagasus,li2023self,liu2023makes}.
Given an IT dataset $D$ containing a sample $S$  = (input $X$, response $Y$),  a designated rating prompt $RP$, and the foundation LLMs $M$, the goal is to leverage both $RP$ and $S$ to prompt $M$ to assign an evaluation $Score$ to the sample $S$ on a scale from 1 to $K$. A higher score typically signifies superior IT data $Quality$.
\begin{equation}
\label{Primary Study}
Quality \propto Score \in [1,K]  = M (RP, S)
\end{equation}

While existing methods~\citep{chen2023alpagasus,cao2023instruction} are adept at identifying high-quality samples, they often over-rely on external resources.
To address these challenges, we introduce SelectIT, a strategy that capitalizes on the internal uncertainty of LLMs to efficiently select high-quality IT data.
SelectIT incorporates three grains of sample evaluation modules: token, sentence, and model-level self-reflections, which effectively improve the reliability of IT data selection.
The comprehensive framework of SelectIT is depicted in Figure~\ref{table:main_pic}.

\subsection{Token-level Self-Reflection}
Numerous studies have demonstrated that foundation models exhibit robust capabilities for next-token prediction during their pre-training phase~\citep{touvron2023LLaMA,touvron2023LLaMA1}.
Yet, this predictive strength is frequently underutilized in evaluating IT data quality. 
In SelectIT, we adopt a similar idea to evaluate IT data.
Specifically, we calculate the next-token probability (from 1 to $K$) based on the rating prompt $RP$ and sample $S$.  
The score token with the highest probability is then considered as the sample's quality.
\begin{equation}
\label{eq1}
S^{base} = \underset{k \in \{1, \ldots, K\}}{\arg\max} \, P'_k ,  P'_k = \left( \frac{{P_k}}{\sum_{j=1}^{K} {P_j}} \right)
\end{equation}
where $P_k$ and $P'_k$ mean the probability and normalized probability of token $k$.

The probability distribution among score tokens reflects the internal uncertainty of LLMs on sample evaluation.
The higher $P'_{S^{base}}$, the more confidence of LLMs, which is not well exploited in Equation~\ref{eq1}.
To capture this subtle difference, we introduce the token-level self-reflection (Token-R), which uses the distribution between tokens that reflect the internal uncertainty of LLMs, to enhance the credibility of quality assessment.
Specifically, we assess the average disparity between the predicted $S^{base}$ token and the other, where the greater the disparity, the more the confidence of LLMs.
This disparity is then utilized to refine the original $S^{base}$, resulting in a token-level score $S^{token}$.
\begin{equation}
\label{token-level score}
S^{token} = S^{base} \times  \underbrace {\frac{1}{K-1} \sum\limits_{i=1}^{K} |P'_i - P'_{S^{base}}|}_{Uncertainty}
\end{equation}

\subsection{Sentence-level Self-Reflection}
Different prompts can significantly affect outputs of LLMs~\citep{kung2023active,peng-etal-2023-towards}, introducing uncertainty into IT data evaluation at the sentence level.
To make better use of this uncertainty to bolster the reliability of our method, we implement sentence-level self-reflection (Sentence-R).
Building upon Token-R, we devise $K$ semantically similar rating prompts$\{RP_0, RP_1, \ldots, RP_K\}$  to obtain a series of quality scores $\{S^{token}_0, S^{token}_1, \ldots, S^{token}_K\}$ based on a given sample $S$.
We calculate the average of these scores to represent the overall quality of sample $S$, because of the importance of incorporating assessments from diverse prompts.
Additionally, we use the standard deviation to quantify the LLMs' uncertainty to rating prompt; a higher standard deviation suggests greater sensitivity to prompt variation, while a lower standard deviation indicates more consistent and confident quality ratings by LLMs~\citep{zhou2020uncertainty}.
By integrating a holistic sample evaluation with the quantification of model uncertainty, we derive the sentence-level score $S^{sent}$, offering a more nuanced and reliable measure of IT data quality.
\begin{equation}
\label{sentence-level score}
S^{sent} =\frac{ \textbf{Avg} \{S^{token}_i\}^K_{i=1}}{1 + \alpha  \times \underbrace {\textbf{Std} \{S^{token}_i\}^K_{i=1}}_{Uncertainty}}
\end{equation}


where ${ \textbf{Avg} \{\cdot\}}$ and ${ \textbf{Std} \{\cdot\}}$ respectively denote the mean and standard deviation of $S^{token}_i$, K means the number of rating prompts $RP$. Moreover, we use the uncertainty factor $\alpha$ to control for the impact of the uncertainty of LLMs on overall scores.

\subsection{Model-level Self-Reflection}

A sample affirmed by multiple foundation models can truly be deemed as high-quality. 
Different foundation models have different quality assessments of the sample, which introduce model-level uncertainty.
To maximize the utilization of this uncertainty, we introduce model-level self-reflection (Model-R).
This strategy leverages the capabilities of existing open-source models without the need for additional resources or the complexities associated with fine-tuning.
However, the challenge lies in the diverse capabilities of various LLMs and determining how to reasonably combine their sample evaluation based on their performance.
It is widely acknowledged that the capabilities of LLMs tend to increase with their parameter count~\citep{hendrycks2020measuring}. 
Thus, we suggest using the parameter count of LLMs as an initial metric for assessing their capabilities to properly weight sample quality scores.
Given $N$ foundation models with parameter counts \(\{\theta_1, \theta_2, \ldots, \theta_N\}\) and their respective sentence-level scores for a sample \(S\) being \(\{S^{sent}_0, S^{sent}_1, \ldots, S^{sent}_N\}\), we formulate the model-level score \(S^{model}\) to reflect a comprehensive evaluation of sample quality.
\begin{equation}
\label{model-level score}
Quality \propto {S^{model}} =\sum_{i=1}^{N} \left( \frac{\theta_i}{\sum_{j=1}^{N} \theta_j} \times S^{sent}_i \right)
\end{equation} 
where $N$ means the number of the foundation models.
By obtaining LLM parameters without resource expenditure, Model-R effectively allows us to employ more powerful foundation models, which is advantageous for selecting higher-quality data.
Finally, we use $S^{model}$ as the final evaluation of sample $S$ in SelectIT. The higher $S^{model}$, the better sample quality. We sort the samples in descending order based on their $S^{model}$ and then select the top-ranked samples as high-quality data.

\subsection{Selective Alpaca}
We apply SelectIT to the widely-used Alpaca-GPT4~\citep{peng2023instruction}.
Specifically, we use the most popular LLaMA-2 (7B, 13B, 70B) as our foundation models and set the hyper-parameters
$\alpha=0.2$ and $K = 5$, which decides the range of LLMs rating in Token-R and the number of rating prompts in Sentence-R.
We finally select the top 20\%, a total of 10.4K pairs as the high-quality data and obtain a curated IT dataset called \textit{Selective Alpaca}.

\section{Experiments}

\subsection{Setups}
\paragraph{Benchmark}
To gain a more comprehensive understanding of the capabilities of LLMs, we evaluate our approach in diverse downstream tasks~\citep{wang2023far,ivison2023camels}.
\textit{Factual knowledge}: We use the Massive Multitask Language Understanding dataset (MMLU~\citep{hendrycks2020measuring}) to assess the factual knowledge of LLMs and report 5-shot results.
\textit{Reasoning}: We evaluate the reasoning abilities of LLMs using two widely utilized datasets: the Grade School Math dataset (GSM~\citep{cobbe2021training}) and Big-Bench-Hard (BBH~\citep{suzgun2022challenging}) with the CoT setting~\citep{wei2022chain}.
\textit{Multilinguality}: we assess this ability by TyDiQA, a multilingual question-answering benchmark that encompasses 11 diverse languages, with the gold-passage setup.
\textit{Coding}: We evaluate this ability using the HumanEval dataset~\citep{chen2021evaluating} and report pass@10 results with a temperature of 0.8.
\textit{Open-ended generation}: We utilize AlpacaEval~\citep{dubois2023Alpacafarm}, which employs GPT-4 to effectively assess model outputs. This can evaluate whether the text produced by LLMs aligns with humans. 

\paragraph{Implementation Details}
We use LLaMA-2 as our testbed.
We fine-tune it for 3 epochs, with a batch size of 128. 
We use Adam with $\beta_1$ = 0.9,  $\beta_2$ = 0.999, and the cosine learning rate scheduler starts from 2e$-$5, and decays to 0.
we opted for a 4096 input length because it can show the best performance of LLMs.
We employ the beam $=4$ for decoding.
We set the temperature parameter to 0.8 and the top$-$p sampling parameter to 0.9 to improve the originality of the output text while ensuring the accuracy and relevance of the content.

\paragraph{Baselines} We compare with the following baselines:
\begin{itemize}
\item \textbf{Alpaca-GPT4}~\citep{peng2023instruction} is a widely-used IT dataset that implements a self-instruct method to autonomously generate instructions by the advanced GPT4.
\item \textbf{LIMA}~\citep{zhou2023lima} primarily consists of 1000 manually crafted high-quality instructional data, which can better stimulate the alignment capability of LLMs.
\item \textbf{AlpaGasus}~\citep{chen2023alpagasus} involves utilizing the robust ChatGPT to score and select data from the original Alpaca-GPT4 dataset. 
\item \textbf{Q2Q}~\citep{li2023quantity} operates by training a precursor model, determining the quality of the IT data based on the two different loss values within this model.
\item \textbf{Instruction Mining}~\citep{cao2023instruction} entails fitting data features and loss values to derive a formula for assessing data quality. 
\end{itemize}

\begin{table*}[t]
\centering
\scalebox{0.75}{
\begin{tabular}{l cc ccccc  ccc  }
\toprule

\multirow{2}{*}{\bf ID ~ System} &  \multicolumn{2}{c}{\textbf{External}} & \multirow{2}{*}{\textbf{MMLU}} & \multirow{2}{*}{\textbf{BBH}}&  \multirow{2}{*}{\textbf{GSM}}& \multirow{2}{*}{\textbf{TydiQA}} &  \multirow{2}{*}{\textbf{CodeX}}&\multirow{2}{*}{\textbf{AE}} & \multicolumn{2}{c}{\textbf{Overall}}  \\
\cmidrule(lr){2-3} \cmidrule(lr){10-11}
& \text{Model}  & \text{Data}  &  &    &  &  &   &  & \text{{AVG}} & \textit{$\Delta$ ($\uparrow$)}\\
\midrule
\multicolumn{1}{l}{ \it Base Model: LLaMA-2-7B} &\multicolumn{6}{c}{\it Implemented Existing  Method} \\ 
~1 ~~~Alpaca-GPT4 &  &   &  46.5   & 38.4    & 15.0    &43.4    &26.8     &34.2 & 34.1 &-  \\
~2 ~~~LIMA  &  \textcolor{red}{\XSolidBrush} & \textcolor{green}{\Checkmark}   & 45.4    & 37.5   & 14.3   & 45.1   &24.6     & 33.1  &33.3  &   \textcolor[RGB]{255,25,0}{-0.7}\\
~3 ~~~1 + AlpaGasus  & \textcolor{green}{\Checkmark} & \textcolor{red}{\XSolidBrush}  & 45.9    & 39.0    & 14.5   &  46.4    &27.5    & 35.4  & 34.8  & \textcolor[RGB]{0,176,80}{+0.7}   \\
~4 ~~~1 + Q2Q  & \textcolor{green}{\Checkmark}  &  \textcolor{red}{\XSolidBrush}   & 46.9   & 39.4    & 15.3   &46.7   &28.2   &35.7   & 35.4   &  \textcolor[RGB]{0,176,80}{+1.3}  \\
~5 ~~~1 + Instruction Mining   & \textcolor{green}{\Checkmark}  &   \textcolor{green}{\Checkmark} & 47.0   & 39.6  & 16.5  &  47.1   &28.6   &34.4  &35.5  & \textcolor[RGB]{0,176,80}{+1.5}   \\
\hdashline 
\multicolumn{10}{c}{\it Our Proposed Method (Individual)}  \\
~6 ~~~1 + Token-R  & \textcolor{red}{\XSolidBrush}&  \textcolor{red}{\XSolidBrush}  & {46.8 }  & 36.5   & 14.5  & 44.6    &28.9   & 35.5  &34.5  & 
\textcolor[RGB]{0,176,80}{+0.4}\\
~7 ~~~1 + Sentence-R  & \textcolor{red}{\XSolidBrush}&  \textcolor{red}{\XSolidBrush}  & 46.9     & 38.1   & 16.1     &\textbf{48.4 } &26.9   & 35.3   &35.3 
& \textcolor[RGB]{0,176,80}{+1.2}\\
~8 ~~~1 + Model-R   & \textcolor{red}{\XSolidBrush}&  \textcolor{red}{\XSolidBrush}  & 47.3   & 37.4    & 16.1   &  45.3     &28.4    & \textbf{35.8 }&35.1   & \textcolor[RGB]{0,176,80}{+1.0}  \\
\hdashline 
\multicolumn{10}{c}{\it Our Proposed Method (All)} \\
~9 ~~~SelectIT (6 + 7 + 8) & \textcolor{red}{\XSolidBrush}&  \textcolor{red}{\XSolidBrush}  & \textbf{47.4}   & \textbf{40.6} & \textbf{16.8}  & 47.4     & \textbf{29.4 }  & 35.7 & \textbf{36.2 } & \textcolor[RGB]{0,176,80}{\textbf{+2.2}}  \\
\midrule
\midrule
\multicolumn{1}{l}{ \it Base Model: LLaMA-2-13B} &\multicolumn{6}{c}{\it Implemented Existing  Method} \\ 

10 ~~~Alpaca-GPT4 & & &  \textbf{55.7}   & 46.6     & 30.5     & 48.1      &40.8      & 46.5   & 44.7    & -  \\
11 ~~~LIMA  & \textcolor{red}{\XSolidBrush}  & \textcolor{green}{\Checkmark}    & 54.6    & 45.3  & 30.5   &  51.1   &34.1   &42.6  &43.0  &   \textcolor[RGB]{255,25,0}{-1.7}  \\

12 ~~10 + AlpaGasus  & \textcolor{green}{\Checkmark}   & \textcolor{red}{\XSolidBrush}   & 54.1    & 47.3       & 31.5     &50.6  &41.3      & 46.3 &45.2   &  \textcolor[RGB]{0,176,80}{+0.5} \\

13 ~~10 + Q2Q & \textcolor{green}{\Checkmark}   &  \textcolor{red}{\XSolidBrush}  & 55.3      & 48.5   & 32.0     &50.8    &41.3     &47.3   &45.9   &  \textcolor[RGB]{0,176,80}{+1.2}   \\

14 ~~10 + Instruction Mining   & \textcolor{green}{\Checkmark}  &   \textcolor{green}{\Checkmark}  & 54.1     &  47.3      &32.5     &52.6    & \textbf{43.3}     & 48.3 &  46.3  &  \textcolor[RGB]{0,176,80}{+1.6}  \\

\hdashline 
\multicolumn{10}{c}{\it Our Proposed Method (Individual)}  \\
15 ~~10 + Token-R  &  \textcolor{red}{\XSolidBrush}&  \textcolor{red}{\XSolidBrush}  & 55.3   & 47.3  & 30.5  & 51.3   &39.8   & 46.2  &45.1  & \textcolor[RGB]{0,176,80}{+0.4}   \\  	 	 	 	 	 	 	 
16 ~~10 + Sentence-R  &  \textcolor{red}{\XSolidBrush}&  \textcolor{red}{\XSolidBrush}  & 55.2    & 48.3    & 31.0    & 52.2    &42.5    & 46.3  &45.9  & \textcolor[RGB]{0,176,80}{+1.2 }  \\
17 ~~10 + Model-R  &   \textcolor{red}{\XSolidBrush}  &  \textcolor{red}{\XSolidBrush} & 55.1   &47.5   & 31.5    & 52.3 &40.2   & 46.1  &45.5  & \textcolor[RGB]{0,176,80}{+0.8}  \\
\hdashline
\multicolumn{10}{c}{\it Our Proposed Method (All)} \\
18 ~~SelectIT (15 + 16 + 17) & \textcolor{red}{\XSolidBrush}&  \textcolor{red}{\XSolidBrush} & \textbf{55.7}     &\textbf{48.9 }   & \textbf{33.0}    & \textbf{54.1}  & {42.2}  & \textbf{48.8} &  \textbf{47.1} &  \textcolor[RGB]{0,176,80}{\textbf{+2.4}}   \\

\bottomrule
\end{tabular}}
\caption{Overall results on IT. ``CodeX'' and ``AE'' mean HumanEval and AlpacaEval benchmarks. All the scores are averages of three independent runs with different random seeds.}
\label{table:main LLMs}
\end{table*}

\subsection{Main Results}
We focus on the discussion of LLaMA-2-13B because both 7B and 13B models exhibit similar trends in Table~\ref{table:main LLMs}.
System (10) shows the vanilla IT on LLMs with the original Alpaca.
By using the data selection strategies, the ability of LLMs has a moderate enhancement in Systems (12) to (14).
Additionally, we can use $S^{base}$ as the input for Equations~\ref{sentence-level score} and~\ref{model-level score} to construct individual methods of Sentence-R and Model-R.
Systems (15) to (17) illustrate that applying each submodule of SelectIT incrementally enhances LLMs' performance, rivaling contemporary advanced methods.

Most remarkably, SelectIT can better boost LLaMA-2's performance compared to vanilla IT in the System (18).
Compared to other IT data selection strategies, this enhancement is particularly evident in the computational and reasoning tasks on the BBH and GSM benchmarks.
This may be attributed to the characteristics of selected data by SelectIT, and we will analyze this phenomenon in a later section.
These gains in reasoning ability also positively impact the coding proficiency of LLMs.
The improvement of LLMs on the TydiQA dataset is also obvious enough, which shows that SelectIT can effectively eliminate similar samples and retain sufficient diversity in multilingual aspects.

\section{Analysis}
This part aims to answer the research questions through the following experiments: How to select high-quality data in SelectIT? (\S \ref{Abalation Study of SelectIT}) Is SelectIT adaptable to various models and domains? (\S \ref{Robustness across Models and Domains}) How about the efficiency of SelectIT? (\S \ref{Efficiency of SelectIT}) What are the advantages of Selective Alpaca?(\S \ref{Insights of Selective Data Curation})

\subsection{Abalation Study of SelectIT}\label{Abalation Study of SelectIT}

\paragraph{Effect of IT Data Quantity}

\begin{wrapfigure}{r}{6cm}
	\centering
  \vspace{-1.4cm}
 \scalebox{0.16}{
	\includegraphics[]{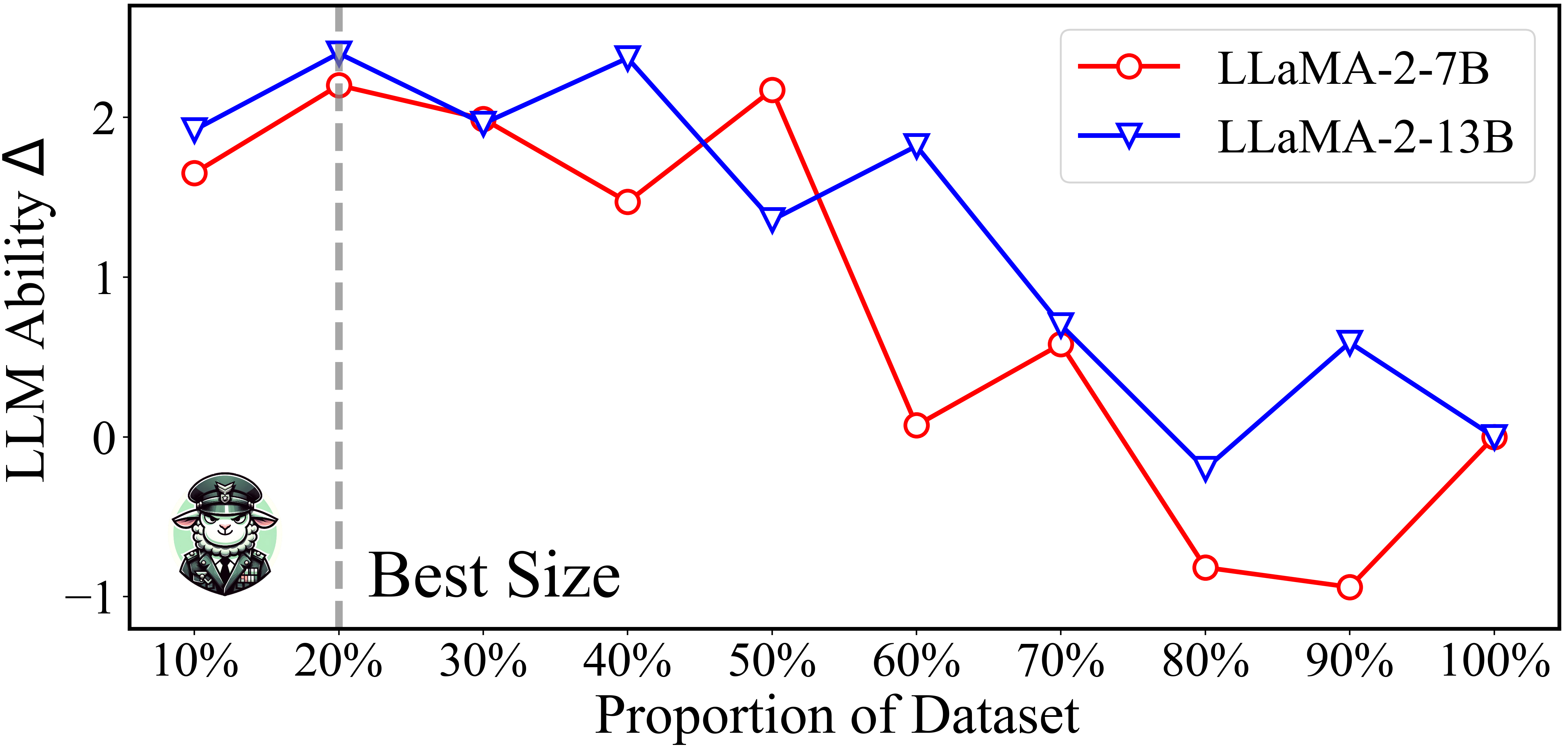}}
	\caption{Comparison of LLM abilities with varying Alpaca proportions.}
	\label{proportion}
 \vspace{-0.7cm}
\end{wrapfigure}
While SelectIT already excels at assessing and ranking samples effectively, selecting an appropriate number of samples in a redundant dataset remains a crucial aspect of our method.
We divide the Alpaca dataset into multiple subsets ranging from 10\% to 100\% based on SelectIT's evaluation and evaluate the overall ability of LLMs on the open-instruct benchmark.
As illustrated in Figure~\ref{proportion}, compared to using the full Alpaca dataset, we observe that LLMs achieve optimal performance using the top 20\% to 40\% data.
Hence, considering the tradeoff of training resources, training time, and model performance, we opt for 20\% for implementing the SelectIT on the Alpaca dataset.

\begin{wraptable}{r}{6cm}
\centering
 \vspace{-0.3cm}
\scalebox{0.75}{
\begin{tabular}{l r r r}
\toprule
{$K$}  & {\bf LLaMA-2-7B}    & {\bf LLaMA-2-13B}  &{\textbf{Overall}}  \\
\midrule
3   &  35.6  & 46.4  & 40.5  \\
5   & \textbf{36.2} & 47.1 & \textbf{41.7}   \\
7   & 35.7  & \textbf{47.3} &41.5  \\
9   & 36.0  & 46.8 & 41.4  \\
\bottomrule 
\end{tabular}
}  
\caption{Effect of different $K$.}
 \vspace{-0.2cm}
\label{table:Effect of K}
\end{wraptable}

\paragraph{Effect of Multiple Rating Prompts}
$K$ is a critical parameter for our method, impacting not only the range of scores assigned by the LLMs but also the number of rating prompts.
We set $K = {3,5,7,9}$ and apply SelectIT for sample selection within the Alpaca to get different subset datasets.
Table~\ref{table:Effect of K} indicates that variations in the value of $K$ have a minor impact on the overall performance of the LLMs. 
This is attributed to our multi-granularity self-reflection mechanism, which effectively enhances the accuracy and stability of sample selection.
Although the model achieves competitive performance at $K$ values of 5 and 7, to minimize resource consumption, we set $K = 5$ as the default value in SelectIT.

\begin{wraptable}{r}{8cm}
\centering
 \vspace{-0.3cm}
\scalebox{0.75}{
\begin{tabular}{l   c  c c c c c c  }
\toprule
 $\alpha$     & {\textbf{MMLU}}  &{\textbf{BBH}}   &{\textbf{GSM}}   & {\textbf{Tydiqa}}& {\textbf{CodeX}} & {\textbf{AE}}  & {\textbf{AVG}}   \\
\midrule
0.2   &  47.4   & \textbf{40.6}    & \textbf{16.8}   & \textbf{47.4}  &\textbf{29.4}  & {35.7}  & \textbf{36.2}   \\

 0.4 & \textbf{47.9}    & {39.4}    &{15.5}     & {46.5}   & \textbf{29.4}  & \textbf{35.8}  & {35.8}   \\

0.6 &  {47.8}  &39.8  & 16.5   & {45.6}  & 29.1    & {35.1}   & 35.7 \\
0.8   & {47.6}     &{36.4}     &{16.5}    & {43.6}   & {26.7}  & {35.4} & {34.4}  \\

\bottomrule

\end{tabular}
}  
\caption{Effect of different $\alpha$.}
 \vspace{-0.22cm}
\label{table:Effect of alpha}
\end{wraptable}

\paragraph{Effect of Uncertainty}
$\alpha$ is an uncertainty factor, integral to calibrating the equilibrium between the mean and the standard deviation of scores derived from Token-R. We assign $\alpha$ four different values, i.e., 0.2, 0.4, 0.6, and 0.8, and incorporate SelectIT for sample selection from within Alpaca to generate disparate subset datasets, with all other parameters remaining constant. 
As shown in Table~\ref{table:Effect of alpha}, with a rise in the $\alpha$ value, Sentece-R tends to emphasize the uncertainty innate to LLMs. This results in the neglect of the average score, a fundamental indicator of sample quality, thereby contributing to a decrease in the overall performance of LLMs. Consequently, we ascertain that an $\alpha$ value of 0.2 is optimally suited to establish an effective balance between the sample's quality and the model's uncertainty.

\begin{wraptable}{r}{6cm}
\centering
 \vspace{-0.4cm}
\scalebox{0.75}{
\begin{tabular}{l l r  r}
\toprule
\textbf{ID}  & \textbf{ Individual }   & \textbf{Unique (\%)}   & \textbf{Overall (\%)}  \\

\midrule
6   &  {Token-R} & 6.18 & 17.83 \\
7   & {Sentence-R} & \textbf{40.81} & \textbf{63.98} \\
8   &  {Model-R}  & 7.37&23.08  \\
\bottomrule 
\end{tabular}
}  
\caption{The relationship between the SelectIT and the individual selection strategy.
Sentence-R plays the most significant impact on the final rating of the IT data. {IDs 6, 7, and 8 correspond to the system of the same IDs in Table \ref{table:main LLMs}}.}
\label{table:rating strategy}
 \vspace{-0.2cm}

\end{wraptable}
\paragraph{Effect of Different Reflection Strategy}
We analyze the relationship between individual selection strategies and SelectIT, from the following two aspects. 
We first account for the number of high-quality data that can only be selected by a unique selection strategy, referred to as unique selection.
Secondly, we calculate which samples in Selective Alpaca can be selected by individual selection strategies in Selective Alpaca, called overall selection.
As shown in Table~\ref{table:rating strategy}, Sentence-R plays the most important role in the final SelectIT strategy.
This is because rating prompts play an important role in sample evaluation and exploiting the effect of different prompts on LLM can effectively better improve the accuracy of sample evaluation than Token-R and Model-R.
Additionally, this phenomenon also aligns with the model's performance reported in Table~\ref{table:main LLMs}, showing the rationality of our proposed uncertainty-aware self-reflection methods.

\begin{table*}[h]
\centering
\scalebox{0.75}{
\begin{tabular}{l c  c c c  c c c c c c c }
\toprule
\multirow{2}{*}{\textbf{Base Model}}  & \multirow{2}{*}{\textbf{Datasets}} & \multirow{2}{*}{\textbf{Data Size}}   & \multirow{2}{*}{\textbf{MMLU}}  &\multirow{2}{*}{\textbf{BBH}}   & \multirow{2}{*}{\textbf{GSM}}   & \multirow{2}{*}{\textbf{Tydiqa}}& \multirow{2}{*}{\textbf{CodeX}} & \multirow{2}{*}{\textbf{AE}}  & \multicolumn{2}{c}{\textbf{Overall}}   \\
\cmidrule(lr){10-11}
&&&&&&&&&\text{{AVG}} & \textit{$\Delta$ ($\uparrow$)}\\
\midrule
\multirow{4}{*}{\bf LLaMA-2-7B } & {LIMA}  & 1K &  45.4   & 37.5 	   & 14.3   & 45.1  & 24.6  & 33.1 & 33.3  & - \\
&{Selective Alpaca} & 1K   & \textbf{46.6}    & \textbf{41.3}    &\textbf{14.5}     & \textbf{46.2}   & \textbf{30.6}  & \textbf{33.8}  & \textbf{35.5} & \textcolor[RGB]{0,176,80}{\textbf{+2.2}}    \\
\cmidrule(lr){2-11}
~ & {AlpaGasus}  & 9K &  {45.9}  &39.0    & 14.5   & \textbf{46.4 }  & 27.5   & 35.4   & 	34.8  & -  \\
&{Selective Alpaca} & 9K   & \textbf{47.2}   & \textbf{41.3}     &\textbf{18.5}     & {47.6}    & \textbf{28.3}   & \textbf{35.4}  & \textbf{36.4} & \textcolor[RGB]{0,176,80}{\textbf{+1.6}} \\
\bottomrule
\end{tabular}
}  
\caption{Results on IT for different datasets with the same number of instances.}
\label{table:same_size}
\end{table*}

\paragraph{Effect of  Data Imbalance}
To eliminate unfair comparison caused by IT data quantity imbalance, we adjust the size of the Selective Alpaca dataset to 1,000 and 9,229 respectively, aligning with the LIMA~\citep{zhou2023lima} and AlpaGasus~\citep{chen2023alpagasus} datasets.
The results in Table \ref{table:same_size} show that, when facing the same amount of data, SelectIT can still demonstrate better performances, which further illustrates its effectiveness.

\subsection{Robustness across Models, Datasets and Domains}\label{Robustness across Models and Domains}

\paragraph{Various Foundation Models}
Although Selective Alpaca achieved impressive improvements in LLaMA-2, applying it to other foundation models remains a challenging task.
To address this, we apply Selective Alpaca on the Mistral-7B and LLaMA-3-8B LLMs and present our results on the open-instruct benchmark alignment with the above test configuration.
As depicted in Table~\ref{table:mistray}, although Selective Alpaca is selected by the LLaMA-2 models, it is also applicable to the Mistral-7B, LLaMA-3-8B and improves their capabilities across various tasks, especially on MMLU, BBH, and GSM benchmarks.
This experiment fully demonstrates the flexibility of SelectIT which does not rely on a specific foundation model for data selection and the universality of Selective Alpaca which can effectively improve the capabilities of different series or scale LLMs.

\begin{table*}[t]
\centering
\scalebox{0.75}{
\begin{tabular}{l c c c  c c c c c c c }
\toprule
\multirow{2}{*}{\textbf{Base Model}}  & \multirow{2}{*}{\textbf{Datasets}}    & \multirow{2}{*}{\textbf{MMLU}}  &\multirow{2}{*}{\textbf{BBH}}   & \multirow{2}{*}{\textbf{GSM}}   & \multirow{2}{*}{\textbf{Tydiqa}}& \multirow{2}{*}{\textbf{CodeX}} & \multirow{2}{*}{\textbf{AE}}  & \multicolumn{2}{c}{\textbf{Overall}}   \\
\cmidrule(lr){9-10}
&&&&&&&&\text{{AVG}} & \textit{$\Delta$ ($\uparrow$)}\\
\midrule
\multirow{2}{*}{\bf LLaMA-2-7B } & {Alpaca-GPT4}   &  46.5   & 38.4    & 15.0   & 43.4  & 26.8  & 34.2  & 34.1  & - \\
&{Selective Alpaca}    & \textbf{47.4 }    & \textbf{40.6 }    &\textbf{16.8 }     & \textbf{47.4 }   & \textbf{29.4  }  & \textbf{35.7 }  & \textbf{36.2} & \textcolor[RGB]{0,176,80}{\textbf{+2.1}}    \\
\bottomrule
\multirow{2}{*}{\bf LLaMA-2-13B } & {Alpaca-GPT4}   &  \textbf{55.7}  &46.6    & 30.5   & 47.1   & 38.8   & 46.5   & 44.2  & -  \\
&{Selective Alpaca}    & 55.3    & \textbf{48.5 }     &\textbf{32.5 }     & \textbf{54.1 }    & \textbf{41.2 }   & \textbf{47.8 }  & \textbf{46.6 } & \textcolor[RGB]{0,176,80}{\textbf{+2.4}}  \\
\bottomrule
\multirow{2}{*}{\bf Mistral-7B } & {Alpaca-GPT4}   &  52.5   & 51.7    & 33.5   & \textbf{51.1}  & 54.7  & 43.1 &47.8 & -  \\
&{Selective Alpaca}    & \textbf{56.9}    & \textbf{53.7}    &\textbf{36.0}   &{49.3}   & \textbf{55.3}   & \textbf{44.3}  & \textbf{49.3 }& \textcolor[RGB]{0,176,80}{\textbf{+1.5}}    \\
\bottomrule
\multirow{2}{*}{\bf LLaMA-3-8B } & {Alpaca-GPT4}   &  59.6  & 52.3   & 34.5  & \textbf{43.1} & 60.2  & \textbf{48.2} &49.7 & -  \\
&{Selective Alpaca}    & \textbf{61.2}  &\textbf{55.0}    & \textbf{37.5}  &41.1   & \textbf{65.4}   & 47.7 & \textbf{51.3}& \textcolor[RGB]{0,176,80}{\textbf{+1.6}}    \\

\bottomrule

\end{tabular}
}  
\caption{Results of IT with various foundation models.}
\label{table:mistray}
\end{table*}

\paragraph{Various Instruction Tuning Datasets}
We further validate the robustness of SelectIT by deploying it on two additional, widely-utilized datasets: WizardLM~\citep{xu2023wizardlm} and Orca-GPT4~\citep{Orca}. WizardLM introduces an innovative method of using LLMs to auto-generate open-domain instructions of varying complexities. This allows for a controlled variation in instructional difficulty and the dataset comprises 143K samples. Orca-GPT4 on the other hand, leverages rich signals from GPT-4 that include explanation traces, step-by-step thought processes, and other multifaceted instructions, all under the guidance of teacher assistance from ChatGPT. Additionally, we maintain consistent hyperparameters, such as $\alpha$ and $K$, choosing LLaMA-2-7B as our base model. We limit the fine-tuning of these datasets to one epoch.
As shown in Table \ref{table:variousITdata}, SelectIT consistently enhances the performance of the model on both the WizardLM and Orca-GPT4 datasets. 
Notably, this augmentative effect is especially pronounced in the computational and reasoning tasks within the BBH and GSM benchmarks. 
In evaluating three separate IT datasets, specifically Alpaca-GPT4, WizardLM, and the more extensive Orca-GPT4, our extensive experimental conclusions validate the broad utility and durability of SelectIT.

\paragraph{Various Domain-specific Tasks}
Machine translation (MT) is a representative domain-specific task of LLMs.
Previous works have already demonstrated significant improvements with LLMs, but they usually use redundant translation IT datasets. This part tests the robustness of SelectIT on the IT dataset of MT.
We select the powerful MT LLM ALMA~\citep{xu2024a} as our backbone model.

\begin{table*}[t]
\centering
\scalebox{0.75}{
\begin{tabular}{l r c c  c c c c c c c }
\toprule
\multirow{2}{*}{\textbf{Datasets}}  & \multirow{2}{*}{\textbf{Data Size}}    & \multirow{2}{*}{\textbf{MMLU}}  &\multirow{2}{*}{\textbf{BBH}}   & \multirow{2}{*}{\textbf{GSM}}   & \multirow{2}{*}{\textbf{Tydiqa}}& \multirow{2}{*}{\textbf{CodeX}} & \multirow{2}{*}{\textbf{AE}}  & \multicolumn{2}{c}{\textbf{Overall}}   \\
\cmidrule(lr){9-10}
&&&&&&&&\text{{AVG}} & \textit{$\Delta$ ($\uparrow$)}\\
\midrule
{WizardLM } & {143K}   &  43.8   & 37.8    & 10.0   & 41.2  & 25.2  & \textbf{35.3}  & 32.2  & - \\
{WizardLM + SelectIT }&{28.6K}    & \textbf{45.1}    & \textbf{40.1}    &\textbf{11.0}     & \textbf{43.1}   & \textbf{27.5}  & {34.7}  & \textbf{33.6} & \textcolor[RGB]{0,176,80}{\textbf{+1.4}}    \\
\bottomrule
{Orca-GPT4} & {1M}   &  {40.1}  &35.6    & 13.0   & \textbf{46.0}  & 23.3    & \textbf{38.1}   & 32.7  & -  \\
{Orca-GPT4 +  SelectIT}  & 0.2M    & \textbf{43.9}     &\textbf{38.7}     & \textbf{16.5}    & {42.0}   & \textbf{27.7}  & {37.4} & \textbf{34.4} &\textcolor[RGB]{0,176,80}{\textbf{+1.7}}  \\
\bottomrule
\end{tabular}
}  
\caption{Results of IT with various IT datasets.}
\label{table:variousITdata}
\end{table*}

We choose the representative language pairs \{German, Chinese\}$\Leftrightarrow$English from WMT’17 to WMT’20 human-written test datasets, and development and test sets from Flores-200, totaling 30K training examples. We used WMT’22 test data  for testing, and finally, 6K high-quality
\begin{wraptable}{r}{6cm}
\centering

\scalebox{0.6}{
\begin{tabular}{l c cc  }
\toprule
\multirow{2}{*}{\bf Method}  & \multirow{2}{*}{\bf Size}&  \multicolumn{2}{c}{\textbf{ALL}} \\
\cmidrule(lr){3-4}
& & COMET & BLEU  \\   
\midrule
\multicolumn{4}{c}{\it SoTA Models}  \\
NLLB~\scriptsize\citep{costa2022no}  & 54B &  78.8& 26.3 \\
GPT-3.5 & - & 85.6& 34.8\\
GPT-4  &-& 85.8 &35.1 \\
\hdashline 
\multicolumn{4}{c}{\it Existing Method}  \\
LLaMA-2~\scriptsize\citep{touvron2023LLaMA} & 7B& 76.5 & 21.1 \\
TIM~\scriptsize\citep{zeng2023tim}& 7B   &  79.1 & 26.4 \\
SWIE~\scriptsize\citep{chen2023improving} &7B  &80.6 & 27.6  \\
BigTranslate~\scriptsize\citep{yang-etal-2023-BigTranslate} & 13B  &78.8 & 21.9\\
Bayling~\scriptsize\citep{bayling} & 13B & 82.0 & 27.8 \\
\hdashline 
\multicolumn{4}{c}{\it Our Implemented Method} \\
ALMA  & 7B  &83.2&29.7\\
\quad w/ SelectIT &7B  &\textbf{83.7}&\textbf{30.5} \\
\hdashline 
ALMA & 13B  &83.7&31.5\\
\quad w/ SelectIT &13B &\textbf{84.2} & \textbf{32.2} \\
\bottomrule
\end{tabular}
 }
\caption{The overall results on MT LLMs. }
\vspace{-0.8cm}
\label{fig:main MT_LLMs}
\end{wraptable}
 examples were selected using SelectIT. 
We utilize both BLEU~\citep{post-2018-call, ott2018scaling} and COMET~\citep{rei-etal-2022-comet} based on the \textit{wmt22-comet-da} model for evaluation. 
We report results for the two language pairs in four directions, using ALL to represent their average. Table~\ref{fig:main MT_LLMs} shows that SelectIT consistently improves ALMA's translation performance.
These results indicate that SelectIT is a versatile and scalable method, effective not only for IT data selection but also for domain-specific tasks like MT. For more detailed analysis and results, please see Appendix \ref{apd:2}.

\subsection{Efficiency of SelectIT}\label{Efficiency of SelectIT}
\begin{wraptable}{r}{6cm}
\centering
\vspace{-0.4cm}
\scalebox{0.75}{
\begin{tabular}{l rr r}
\toprule
{\textbf{Method}} & {\textbf{Speed}} & {\textbf{Time}} & {\textbf{Cost}} \\
\midrule
ChatGPT API  &  0.76 it/s  & 19.07h  & \$52.02  \\
GPT4 API   & 0.37 it/s & 38.98h & \$2871.56 \\
SelectIT   & \textbf{9.34 it/s}  & \textbf{5.80h} & \textbf{\$26.68} \\
\bottomrule 
\end{tabular}
}  
\caption{Comparison of selection efficiency.}
\vspace{-0.22cm}
\label{table:Efficiency}
\end{wraptable}
SelectIT is a faster and more cost-effective method for IT data selection. We compared different selection methods on the Alpaca-GPT4 dataset. 
For ChatGPT (AlpaGasus) or GPT-4, we randomly select 500 instruction data from Alpaca-GPT4, analyze various metrics, and estimate the resource consumption for selecting the entire dataset.
Using SelectIT, we employ 4 A800 80G GPUs to select high-quality IT data, calculating the total cost based on Google Cloud's rate of \$1.15/h per single GPU.
As shown in Table \ref{table:Efficiency}, SelectIT is significantly faster and uses the least resources. 
This efficiency is due to computing only the probability of the next token for input sentences, bypassing the full sentence generation and decoding process, resulting in lower resource consumption.
Additionally, using our own GPU at a low cost enhances transparency, allowing us to preserve all intermediate outputs and results for thorough analysis in data selection.

\subsection{Insights of Selective Data Curation}\label{Insights of Selective Data Curation}
\paragraph{Different Selection Strategies}
\begin{wraptable}{r}{6cm}
\vspace{-0.4 cm}
\centering
\scalebox{0.6}{
\begin{tabular}{l cc cc c  }
\toprule
\multirow{2}{*}{Method}  & \multicolumn{2}{c}{\bf LLaMA-2}    & \multicolumn{2}{c}{\bf ALMA } & \multirow{2}{*}{\textit{$\Delta$ ($\uparrow$)} } \\
\cmidrule(lr){2-3} \cmidrule(lr){4-5}
&7B&13B&7B&13B& \\
\midrule
Full Dataset  &  34.1 & 44.2  & 29.7 & 31.5 & -\\
\quad w/ Random (Full)   & 34.1 & 45.1 & 29.3 & 31.0 &{0.0}  \\
\quad w/ Random (Unselected)    & 34.6& 44.3  & 29.1 &31.2 & \textcolor[RGB]{255,25,0}{-0.4}  \\

\quad w/ Length   & 35.5 & 47.1 & 30.1 &31.8  & \textcolor[RGB]{0,176,80}{+5.0}  \\

\quad w/ SelectIT   & \textbf{36.2}  & \textbf{47.1} & \textbf{30.5} & \textbf{32.2} &  \textcolor[RGB]{0,176,80}{\textbf{+6.5}}\\
\bottomrule 
\end{tabular}
}  
\caption{Comparasion with variants.}
\vspace{-0.3cm}
\label{table:Random Selection}
\end{wraptable}

This part compares three different selection strategies, namely, randomly selecting 20\% in the full Alpaca and unselected dataset of Selective Alpaca, and selecting 20\% data based on sample length~\citep{zhao2024long}.
As shown in Table~\ref{table:Random Selection}, the random-based strategies show certain performance degradation and the random selection in the unselected dataset is even worse, which reflects the effectiveness of our method from the side.
Selection based on sample length is a simple approach to defining high-quality data, but it does not take into account the content of IT data, resulting in the limited performance of LLMs.
SelectIT can significantly improve the abilities of LLMs.

\begin{figure*}[t]
    \centering
\begin{minipage}{0.5\linewidth}
    \centering
    \includegraphics[width=\linewidth]{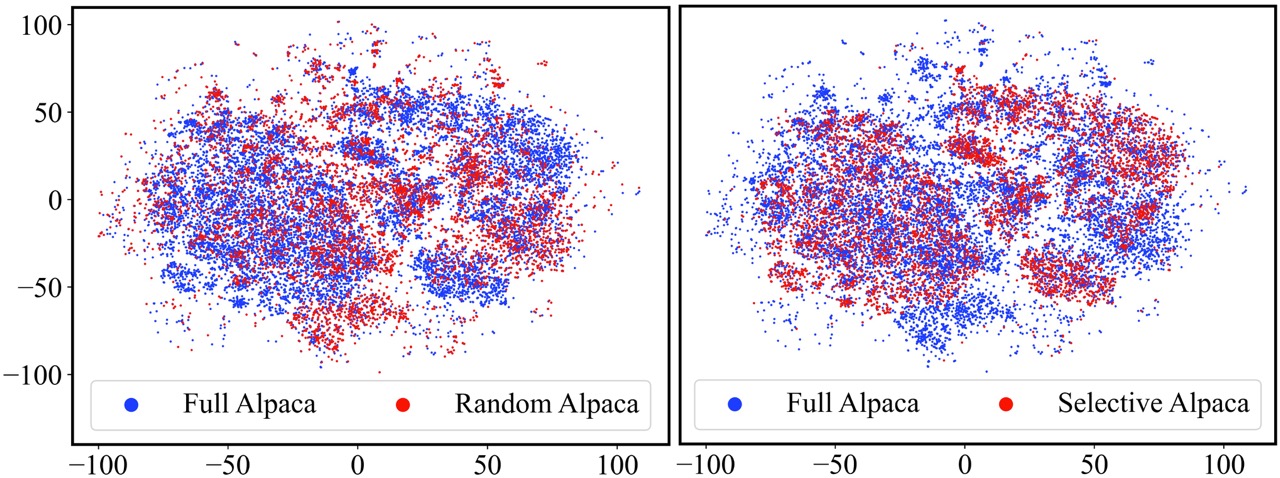}
	\caption{Instruction embeddings representations of different selection strategies. The red and blue points are representations of full Alpaca datasets and selected data respectively.}
    \label{tsne}
\end{minipage}
\hspace{0.3em} 
\begin{minipage}{0.45\linewidth}
    \includegraphics[width=\linewidth]{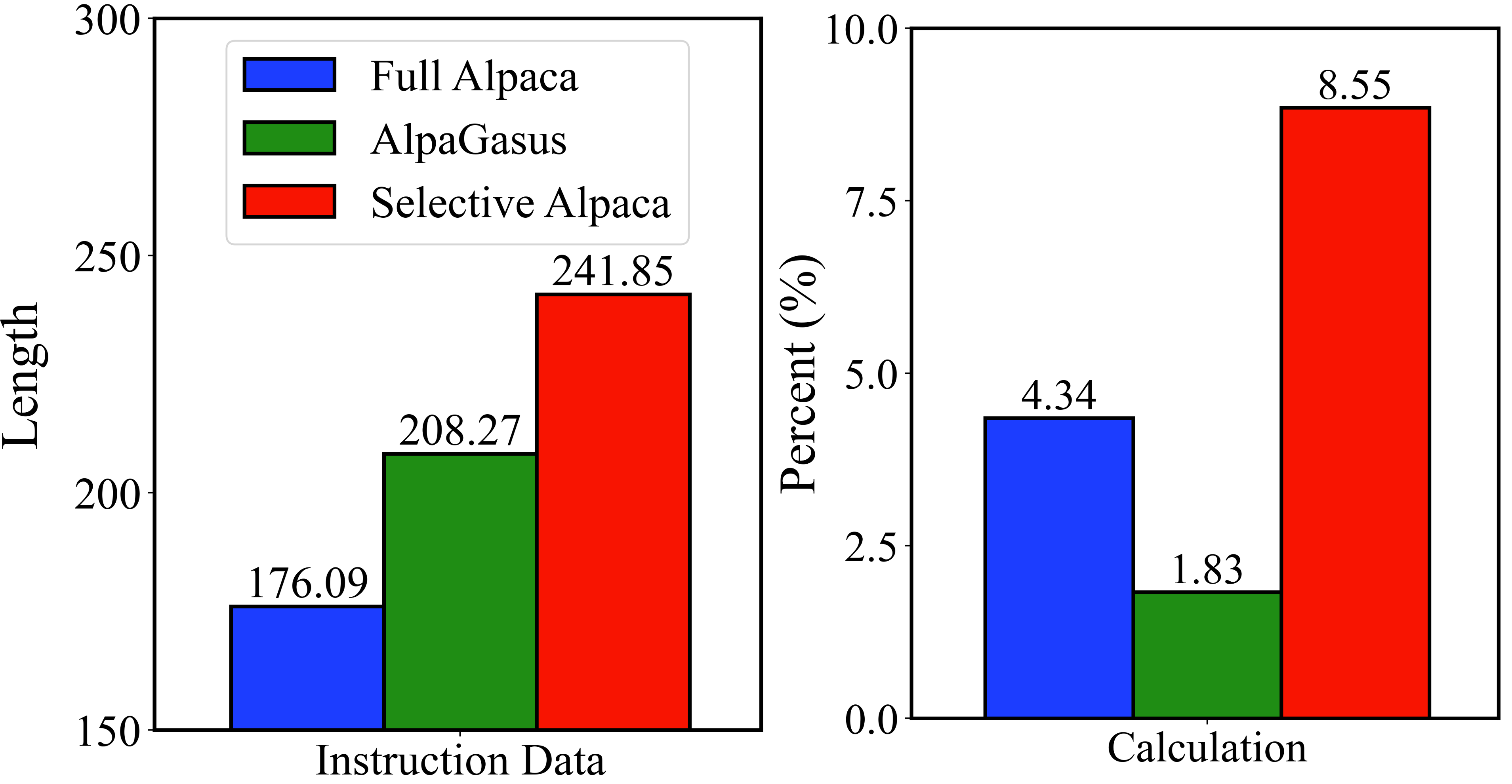}
	\caption{Left: The average length of samples. Right: The proportion of calculation type.}
    \label{data_analysis}
\end{minipage}
\vspace{-0.3cm}
\end{figure*}

\paragraph{Data Representation Analysis}
This part explores the relationship between Selective Alpaca and the original datasets from a representation perspective. Following \citet{gao2024towards}, we use the outputs of the last layer corresponding to the last token in the input sequence as sample representations. We then apply T-SNE \citep{hinton2002stochastic} for dimensionality reduction, mapping high-dimensional embeddings onto a 2D space.
Figure~\ref{tsne} shows the intermediate representations generated by the full and Selective Alpaca datasets. Randomly selected data struggle to distinguish abnormal data far from the center, making it hard to define high-quality IT data. In contrast, Selective Alpaca data are mostly concentrated around the center, indicating that our dataset predominantly contains high-quality data near the center and effectively discards abnormal data, supporting the conclusion of Table \ref{table:Random Selection}.

\paragraph{Data Characteristic Analysis}
We analyze the Selective Alpaca from the following two perspectives, to explore why our dataset is better than the original dataset and its variants.
Firstly, as shown in Figure~\ref{data_analysis}, the length of instructions from the Selective Alpaca is significantly longer than those in the Alpaca dataset and AlpaGasus which is selected by ChatGPT.
This implies that, with the same amount of data, our dataset contains more information, aligned with the results in Table~\ref{table:Random Selection}.
Secondly, by using ChatGPT to examine IT data types, we find a substantial increase in the proportion of computational problems in Selective Alpaca.
This indicates that Selective Alpaca tends to select high-quality mathematical data, providing a solid explanation for the observed improvement in the reasoning abilities of LLMs as demonstrated in Table~\ref{table:main LLMs}.
Appendix~\ref{apd:1} shows the case study of comparing the Selective Alpaca with AlpaGasus.

\begin{wrapfigure}{r}{6cm}
\centering
\vspace{-0.4cm}
 \scalebox{0.18}{
	\includegraphics[]{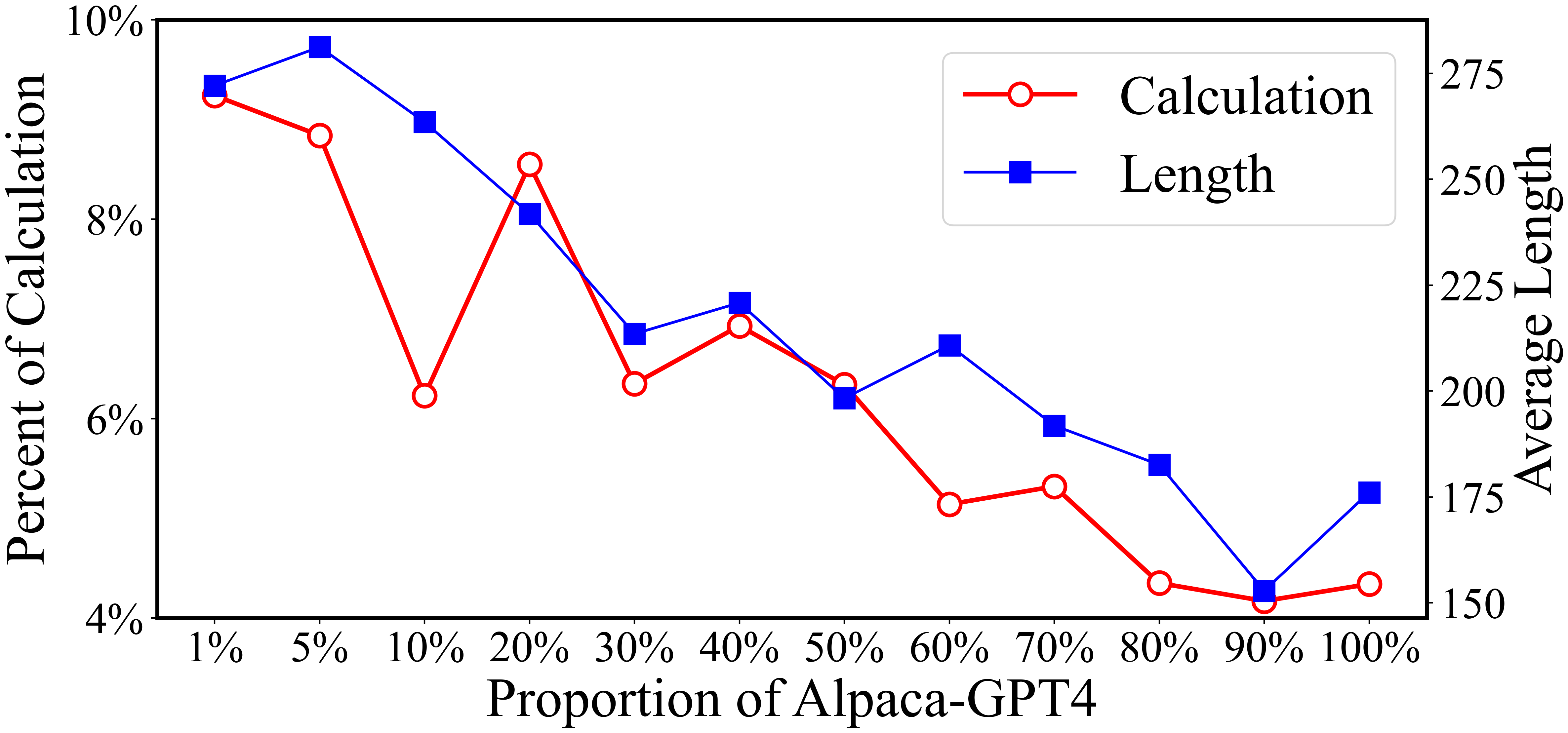}}
	\caption{Changing trends of the calculation and sample length with different data sizes.}
	\label{proportion_calculation}
 \vspace{-0.4cm}
\end{wrapfigure}
\paragraph{Insights of High-Quality Data in SelectIT}
Furthermore, we analyze the proportion of calculation and sample average length in Alpaca-GPT4 with different proportions after sorting by SelectIT to explore its intrinsic characteristics and the definition of high-quality data.
As shown in Figure \ref{proportion_calculation}, with the proportion of Alpaca-GPT4 data continuing to increase, the proportion of calculation and sample average length gradually decreases.
This phenomenon clearly indicates that SelectIT can reasonably rank samples based on their characteristics.
When the data size is more than 50\%, the proportion of calculation IT data sharply declines, falling below 6\%, causing a noticeable decrease in the model's overall capability, as depicted in Figure \ref{proportion}. 
This analysis shows that more computationally intensive IT data may be a new perspective on the characteristics of optimal IT data, which not only effectively improves the LLMs' reasoning ability, but also further drives the improvement of other abilities.

\section{Conclusion}
This paper introduces a novel data selection strategy, SelectIT, for LLM instruction tuning, which uses LLM uncertainty to efficiently identify high-quality IT data without requiring additional resources.
SelectIT includes three types of self-reflection: token, sentence, and model, which can individually and jointly improve the performance of IT data selection.
By applying SelectIT to the Alpaca-GPT4 dataset, we introduce a compact and strong IT dataset, called Selective Alpaca. Different models and domain tasks demonstrate the effectiveness of SelectIT.
Our analysis reveals that SelectIT effectively excludes abnormal data and tends to select longer and calculational data.

\section*{Limitation}
{This paper could be further strengthened as follows:
}\begin{itemize}
    \item \textbf{Instruction Data Quantity}: {Our findings suggest that prioritizing the top 20\% of high-quality data optimizes results for Alpaca. Future studies might explore adjusting this threshold based on the data quality in different datasets to enhance performance.}
    \item \textbf{Models at Different Scales}: {Our analysis is currently limited to models smaller than 30B parameters due to computational constraints. Investigating the efficacy of Selective Alpaca on larger-scale LLMs, could provide valuable insights into the method's scalability.}
    \item \textbf{Expansion to Additional Instruction Datasets}: {Although SelectIT has been applied to the Alpaca dataset due to its widespread adoption, extending this methodology to incorporate other IT datasets could offer substantial advantages to the broader LLM research community.}
\end{itemize}

\section*{Broader Impacts}
{Our work follows the NeurIPS Ethics Policy. Our findings are based on publicly available datasets for reproducibility purposes. 
LLMs can contain potential racial and gender bias. Therefore, if someone finds our work interesting and would like to use it in a specific environment, we strongly suggest the user check the potential bias before usage. In addition, it is hard to control the generation of LLMs. We should be aware of the potential problems caused by hallucinations.}

\section*{Acknowledgments}
{This work was supported in part by the National Natural Science Foundation of China (Grant No. 62206076), Guangdong Basic and Applied Basic Research Foundation (Grant No. 2024A1515011491), Shenzhen Science and Technology Program (Grant Nos. ZDSYS20230626091203008, KJZD20231023094700001, RCBS20221008093121053), and Shenzhen College Stability Support Plan (Grant Nos. GXWD20220811173340003, GXWD20220817123150002). Derek F. Wong was supported in part by the Science and Technology Development Fund, Macau SAR (Grant Nos. FDCT/060/2022/AFJ, FDCT/0070/2022/AMJ), National Natural Science Foundation of China (Grant No. 62261160648), the Multi-year Research Grant from the University of Macau (Grant No. MYRG-GRG2024-00165-FST), and the Tencent AI Lab Rhino-Bird Gift Fund (Grant No. EF2023-00151-FST).
We would like to thank the anonymous reviewers and meta-reviewer for their insightful suggestions.}

\bibliography{neurips_2024}

\bibliographystyle{mybst}



\clearpage
\appendix
\section{Appendix}

\subsection{Applying SelectIT on Machine Translation LLMs}
Machine Translation (MT) is a important task for LLMs, demonstrating their domain-specific capabilities. Prior research, including TIM~\citep{zeng2023tim}, SWIE~\citep{chen2023improving}, BigTranslate~\citep{yang-etal-2023-BigTranslate}, and Bayling~\citep{bayling}, has shown significant improvements in LLMs, often relying on extensive translation training datasets. 
In this section, we examine the impact of training data quality on MT performance, employing the robust MT LLM, ALMA, as our foundational model~\citep{xu2024a}.

For training data, we select representative language pairs: German$\Leftrightarrow$English and Chinese$\Leftrightarrow$English, sourced from WMT’17 to WMT’20 human-authored test datasets, supplemented with development and test sets from Flores-200, totaling 30K training instances. We use the corresponding language pair's test data from WMT’22 as evaluation datasets. Subsequently, 6K high-quality instances are selected for LORA fine-tuning via SelectIT.

We report both the widely used BLEU score~\citep{post-2018-call, ott2018scaling} and the COMET score~\citep{rei-etal-2022-comet} based on the \textit{wmt22-comet-da} model, which shows higher correlation with human judgments for evaluating the LLMs' translation abilities. 
Table~\ref{fig:main MT_LLMs_apd} consistently demonstrates that SelectIT enhances ALMA's translation efficacy. 
Notably, SelectIT primarily focuses on improving translations from English to other languages, likely due to ALMA's inherent proficiency in English, which presents challenges for further enhancements. 
These findings highlight SelectIT's adaptability and scalability, validating its effectiveness not only in IT data selection but also in domain-specific tasks such as MT.

\label{apd:2}
\begin{table*}[h]
\centering
\scalebox{0.75}{
\begin{tabular}{l c cc  cc cc  cc cc  }
\toprule
\multirow{2}{*}{\bf Method}  & \multirow{2}{*}{\bf Size}& \multicolumn{2}{c}{\bf En$\Rightarrow$De} & \multicolumn{2}{c}{\bf De$\Rightarrow$En} & \multicolumn{2}{c}{\bf Zh$\Rightarrow$En} & \multicolumn{2}{c}{\bf En$\Rightarrow$Zh} &  \multicolumn{2}{c}{\textbf{ALL}} \\
\cmidrule(lr){3-4}\cmidrule(lr){5-6}\cmidrule(lr){7-8}\cmidrule(lr){9-10}\cmidrule(lr){11-12}
& & COMET & BLEU & COMET & BLEU &COMET & BLEU & COMET & BLEU & COMET & BLEU  \\
\midrule
\multicolumn{11}{c}{\it SoTA Models}  \\
NLLB & 54B& 86.5 &  34.5 &  78.9 & 26.9 & 70.7 & 16.6 & 78.9  & 27.4 &  78.8& 26.3 \\

GPT-3.5 & - &  87.0 &  34.4  &85.5 & 33.1 & 82.9 & 26.6 &87.0 &44.9 &85.6& 34.8\\
GPT-4  &-& 87.4  & 35.4 & 85.6 & 33.9 &82.8& 27.2&87.5& 44.0&85.8 &35.1 \\
\hdashline 
\multicolumn{11}{c}{\it Existing Method}  \\
LLaMA-2 & 7B&  76.4 & 19.0 & 82.7 &  30.4& 75.0  & 18.2  & 71.8 & 17.0 &76.5 & 21.1 \\
TIM& 7B   & 74.2  &20.6   & 77.7 &  24.3  & 79.5 &  23.4 &84.9 &37.2 &  79.1 & 26.4 \\
SWIE &7B  & 82.4 & 27.2 & 83.0  & 30.5 & 76.5 &  21.3  &80.6  & 31.2&80.6 & 27.6  \\
BigTranslate & 13B  & 78.8&  21.5 & 80.7  & 23.4  & 74.3  & 14.2 &81.3  & 28.6&78.8 & 21.9\\
Bayling & 13B & 82.7&  25.6 & 83.0 &  27.3  & 77.7 & 20.1 &84.6   & 37.9 & 82.0 & 27.8 \\
\hdashline 
\multicolumn{11}{c}{\it Our Implemented Method} \\
ALMA & 7B  & 85.0 &  29.9& 83.9 & 30.0 &  79.2&  22.7 & 84.8 & 36.3&83.2&29.7\\
\quad w/ SelectIT &7B  & \textbf{85.2}   &\textbf{30.2 }& \textbf{84.1} & \textbf{~~30.4}$^\dagger$ & \textbf{~~80.0}$^\dagger$  &  \textbf{~~24.2} $^\dagger$&\textbf{~~85.3}$^\dagger$ & \textbf{~~37.3}$^\dagger$&\textbf{83.7}&\textbf{30.5} \\
\hdashline  
ALMA & 13B  &85.2 &31.0&84.2 &30.9 &80.0 &25.0 &85.5& 39.2&83.7&31.5\\
\quad w/ SelectIT &13B  & \textbf{~~85.8}$^\dagger$   & \textbf{~~31.7}$^\dagger$& \textbf{84.6} & \textbf{~~31.4}$^\dagger$& \textbf{80.3}    &  \textbf{25.4} &\textbf{~~86.1}$^\dagger$  &\textbf{~~40.4}$^\dagger$ &\textbf{84.2} & \textbf{32.2} \\
\bottomrule
\end{tabular}
 }
\caption{Overall results on machine translation LLMs. ``$^\dagger$'' the improvement is significant by contrast to the ALMA model ($p$ < 0.05).}
\label{fig:main MT_LLMs_apd}
\end{table*}

 \begin{figure*}[t]
	\centering
 \scalebox{0.25}{
	\includegraphics[]{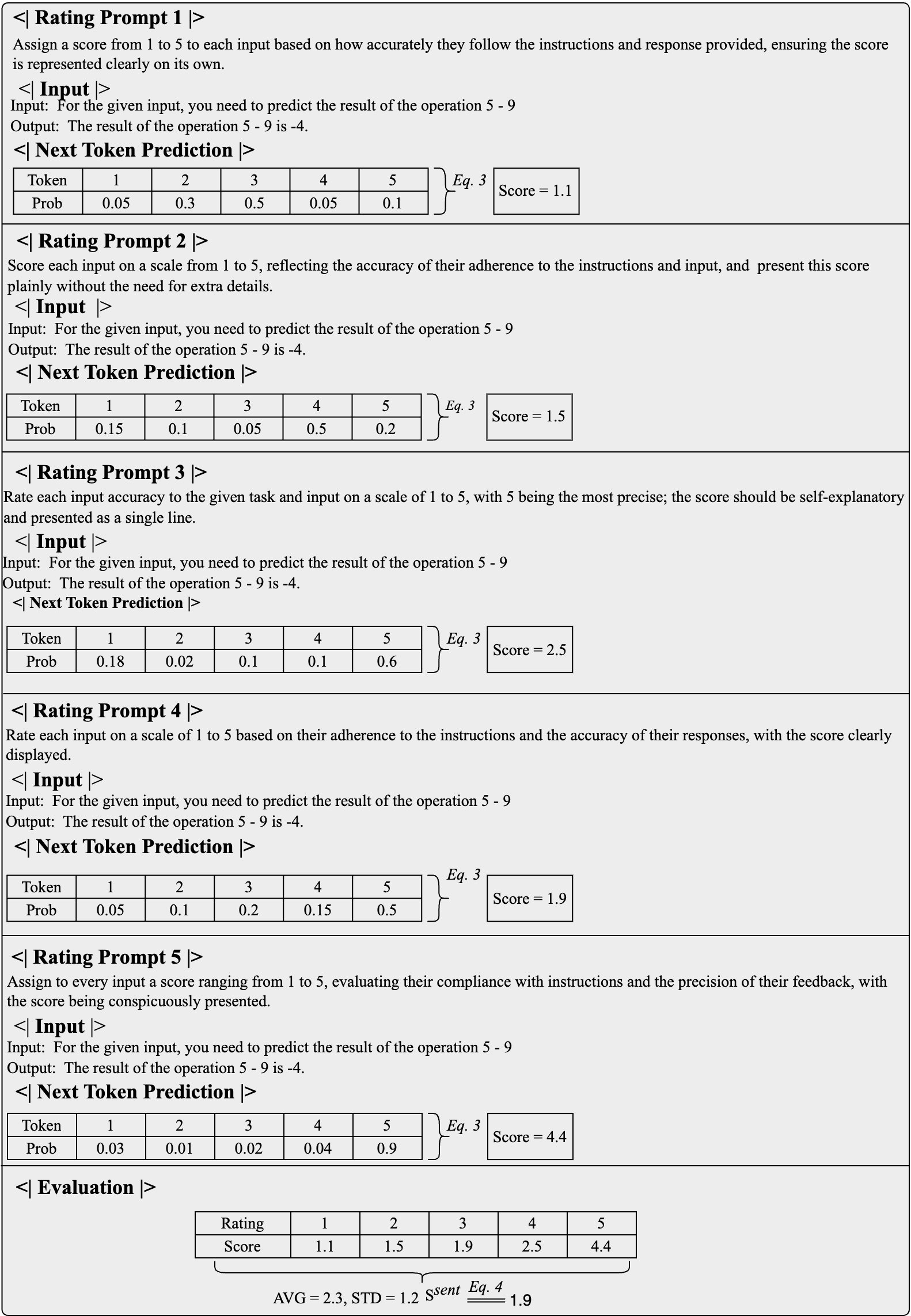}}
	\caption{Example on Sentence-R calculation of SelectIT.}
	\label{table:apd_sent_0}
\end{figure*}

\subsection{Details of Sentence-level Rating }
Based on the preceding analysis, Sentence-R is integral to the functionality of SelectIT. 
As illustrated in Equation~\ref{sentence-level score}, the Token-level Rating forms the foundation for the Sentence-level Rating. 
The Model-level Rating is derived through multiple iterations of the Sentence-level Rating across different foundational LLMs. 
Therefore, a detailed explanation of Sentence-R is sufficient to demonstrate the operational mechanism of SelectIT. 
As depicted in Figure~\ref{table:apd_sent_0}, we utilize five distinct rating prompts along with a single input to formulate the final input for Sentence-R. 
Initially, each rating prompt produces a score of  $S^{token}$. We then compute the mean and standard deviation of these $S^{token}$ values to obtain the final $S^{sent}$, as outlined in Equation~\ref{sentence-level score}.






\subsection{Case Study}
\label{apd:1}
\begin{figure*}[t]
	\centering
 \scalebox{1.2}{
	\includegraphics[]{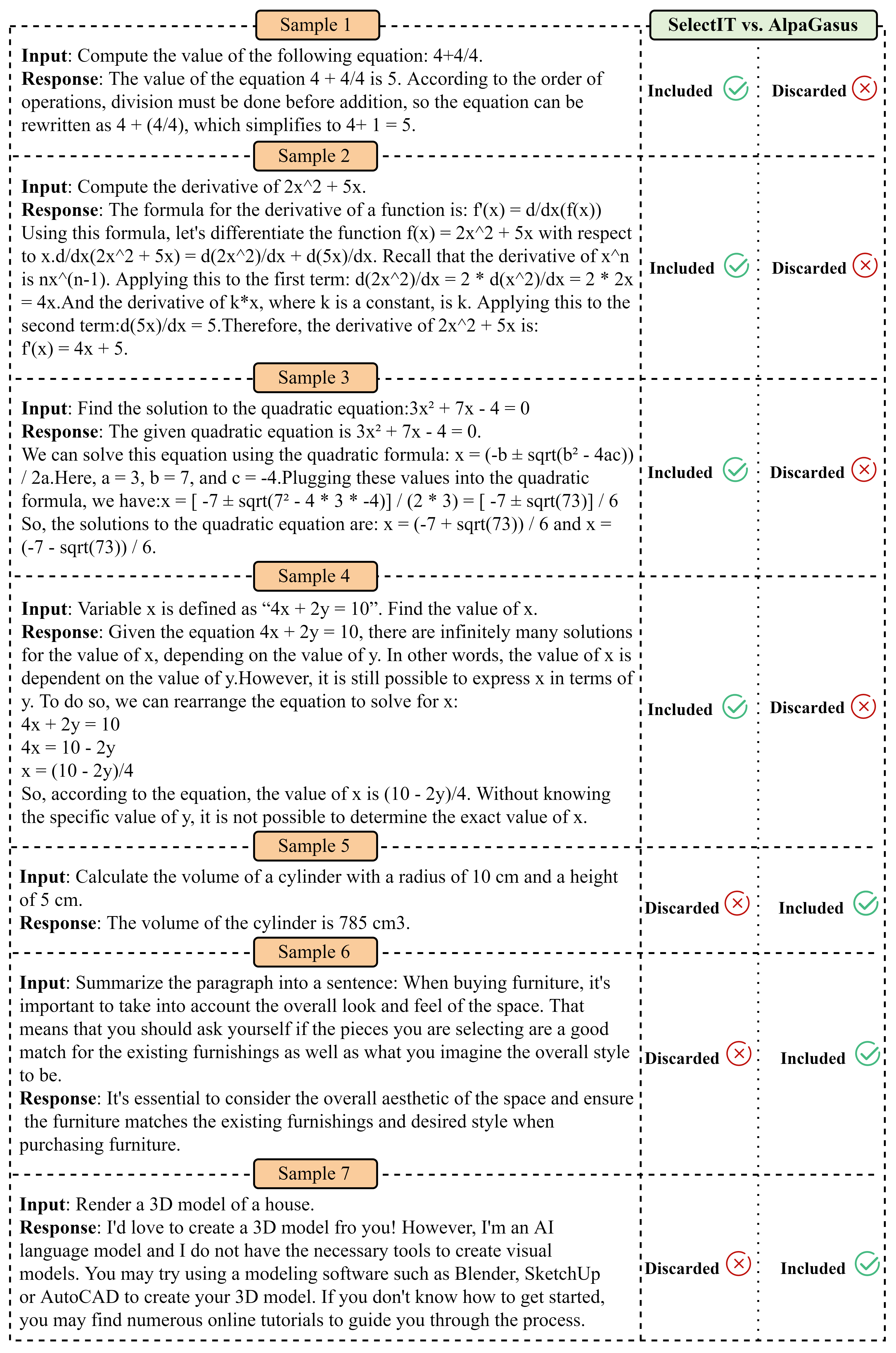}}
	\caption{Examples of IT data selected by SelectIT or AlpaGasus.}
	\label{table:case_study}
\end{figure*}

As demonstrated in Figure~\ref{table:case_study}, we illustrate the selection tendencies of SelectIT in contrast to AlpaGasus, which leverages advanced ChatGPT for data selection. In samples 1 to 4, SelectIT shows a preference for instruction-tuning data containing intricate mathematical problems that contribute to improving the reasoning skills of the LLMs. On the contrary, AlpacaGasus frequently chooses IT data in samples 5 to 7 that primarily offer solutions to queries or lack coherent reasoning, which might limit its effectiveness.

\clearpage
\section*{NeurIPS Paper Checklist}



\begin{enumerate}

\item {\bf Claims}
    \item[] Question: Do the main claims made in the abstract and introduction accurately reflect the paper's contributions and scope?
    \item[] Answer: \answerYes{} 
    \item[] Justification: {See the abstract and introduction sections. Our proposed SelectIT can capitalize on the foundational capabilities of the LLM itself to more effectively select high-quality IT data, without the need for extra resources. We run comprehensive experiments to support our assumption. Our contributions are stated clearly in the abstract and introduction.}
    \item[] Guidelines: 
    \begin{itemize}
        \item The answer NA means that the abstract and introduction do not include the claims made in the paper.
        \item The abstract and/or introduction should clearly state the claims made, including the contributions made in the paper and important assumptions and limitations. A No or NA answer to this question will not be perceived well by the reviewers. 
        \item The claims made should match theoretical and experimental results, and reflect how much the results can be expected to generalize to other settings. 
        \item It is fine to include aspirational goals as motivation as long as it is clear that these goals are not attained by the paper. 
    \end{itemize}

\item {\bf Limitations}
    \item[] Question: Does the paper discuss the limitations of the work performed by the authors?
    \item[] Answer: \answerYes{} 
    \item[] Justification: In the limitation part, we have discussed the points where SelectIT could be further optimized, including the data quantity, model scales, other foundation models, and datasets. 
    \item[] Guidelines:
    \begin{itemize}
        \item The answer NA means that the paper has no limitation while the answer No means that the paper has limitations, but those are not discussed in the paper. 
        \item The authors are encouraged to create a separate "Limitations" section in their paper.
        \item The paper should point out any strong assumptions and how robust the results are to violations of these assumptions (e.g., independence assumptions, noiseless settings, model well-specification, asymptotic approximations only holding locally). The authors should reflect on how these assumptions might be violated in practice and what the implications would be.
        \item The authors should reflect on the scope of the claims made, e.g., if the approach was only tested on a few datasets or with a few runs. In general, empirical results often depend on implicit assumptions, which should be articulated.
        \item The authors should reflect on the factors that influence the performance of the approach. For example, a facial recognition algorithm may perform poorly when image resolution is low or images are taken in low lighting. Or a speech-to-text system might not be used reliably to provide closed captions for online lectures because it fails to handle technical jargon.
        \item The authors should discuss the computational efficiency of the proposed algorithms and how they scale with dataset size.
        \item If applicable, the authors should discuss possible limitations of their approach to address problems of privacy and fairness.
        \item While the authors might fear that complete honesty about limitations might be used by reviewers as grounds for rejection, a worse outcome might be that reviewers discover limitations that aren't acknowledged in the paper. The authors should use their best judgment and recognize that individual actions in favor of transparency play an important role in developing norms that preserve the integrity of the community. Reviewers will be specifically instructed to not penalize honesty concerning limitations.
    \end{itemize}

\item {\bf Theory Assumptions and Proofs}
    \item[] Question: For each theoretical result, does the paper provide the full set of assumptions and a complete (and correct) proof?
    \item[] Answer: \answerNA{} 
    \item[] Justification: Our paper does not include theoretical results.
    \item[] Guidelines:
    \begin{itemize}
        \item The answer NA means that the paper does not include theoretical results. 
        \item All the theorems, formulas, and proofs in the paper should be numbered and cross-referenced.
        \item All assumptions should be clearly stated or referenced in the statement of any theorems.
        \item The proofs can either appear in the main paper or the supplemental material, but if they appear in the supplemental material, the authors are encouraged to provide a short proof sketch to provide intuition. 
        \item Inversely, any informal proof provided in the core of the paper should be complemented by formal proofs provided in appendix or supplemental material.
        \item Theorems and Lemmas that the proof relies upon should be properly referenced. 
    \end{itemize}

    \item {\bf Experimental Result Reproducibility}
    \item[] Question: Does the paper fully disclose all the information needed to reproduce the main experimental results of the paper to the extent that it affects the main claims and/or conclusions of the paper (regardless of whether the code and data are provided or not)?
    \item[] Answer: \answerYes{} 
    \item[] Justification: In the abstract section, we provide the GitHub link to open source all the code, scripts, and datasets (Selective Alpaca) for other researchers to replicate the results. We also provide the implementation details to better reproduce the experiments.
    \item[] Guidelines:
    \begin{itemize}
        \item The answer NA means that the paper does not include experiments.
        \item If the paper includes experiments, a No answer to this question will not be perceived well by the reviewers: Making the paper reproducible is important, regardless of whether the code and data are provided or not.
        \item If the contribution is a dataset and/or model, the authors should describe the steps taken to make their results reproducible or verifiable. 
        \item Depending on the contribution, reproducibility can be accomplished in various ways. For example, if the contribution is a novel architecture, describing the architecture fully might suffice, or if the contribution is a specific model and empirical evaluation, it may be necessary to either make it possible for others to replicate the model with the same dataset, or provide access to the model. In general. releasing code and data is often one good way to accomplish this, but reproducibility can also be provided via detailed instructions for how to replicate the results, access to a hosted model (e.g., in the case of a large language model), releasing of a model checkpoint, or other means that are appropriate to the research performed.
        \item While NeurIPS does not require releasing code, the conference does require all submissions to provide some reasonable avenue for reproducibility, which may depend on the nature of the contribution. For example
        \begin{enumerate}
            \item If the contribution is primarily a new algorithm, the paper should make it clear how to reproduce that algorithm.
            \item If the contribution is primarily a new model architecture, the paper should describe the architecture clearly and fully.
            \item If the contribution is a new model (e.g., a large language model), then there should either be a way to access this model for reproducing the results or a way to reproduce the model (e.g., with an open-source dataset or instructions for how to construct the dataset).
            \item We recognize that reproducibility may be tricky in some cases, in which case authors are welcome to describe the particular way they provide for reproducibility. In the case of closed-source models, it may be that access to the model is limited in some way (e.g., to registered users), but it should be possible for other researchers to have some path to reproducing or verifying the results.
        \end{enumerate}
    \end{itemize}

\item {\bf Open access to data and code}
    \item[] Question: Does the paper provide open access to the data and code, with sufficient instructions to faithfully reproduce the main experimental results, as described in supplemental material?
    \item[] Answer: \answerYes{} 
    \item[] Justification: In the abstract section, we provide the GitHub link to open access to data and code. 
    \item[] Guidelines:
    \begin{itemize}
        \item The answer NA means that paper does not include experiments requiring code.
        \item Please see the NeurIPS code and data submission guidelines (\url{https://nips.cc/public/guides/CodeSubmissionPolicy}) for more details.
        \item While we encourage the release of code and data, we understand that this might not be possible, so “No” is an acceptable answer. Papers cannot be rejected simply for not including code, unless this is central to the contribution (e.g., for a new open-source benchmark).
        \item The instructions should contain the exact command and environment needed to run to reproduce the results. See the NeurIPS code and data submission guidelines (\url{https://nips.cc/public/guides/CodeSubmissionPolicy}) for more details.
        \item The authors should provide instructions on data access and preparation, including how to access the raw data, preprocessed data, intermediate data, and generated data, etc.
        \item The authors should provide scripts to reproduce all experimental results for the new proposed method and baselines. If only a subset of experiments are reproducible, they should state which ones are omitted from the script and why.
        \item At submission time, to preserve anonymity, the authors should release anonymized versions (if applicable).
        \item Providing as much information as possible in supplemental material (appended to the paper) is recommended, but including URLs to data and code is permitted.
    \end{itemize}

\item {\bf Experimental Setting/Details}
    \item[] Question: Does the paper specify all the training and test details (e.g., data splits, hyperparameters, how they were chosen, type of optimizer, etc.) necessary to understand the results?
    \item[] Answer: \answerYes{} 
    \item[] Justification: In Section 4.1, we have discussed datasets, baselines, and experimental setup used in our experiments. More training details are included in the code.
    \item[] Guidelines:
    \begin{itemize}
        \item The answer NA means that the paper does not include experiments.
        \item The experimental setting should be presented in the core of the paper to a level of detail that is necessary to appreciate the results and make sense of them.
        \item The full details can be provided either with the code, in appendix, or as supplemental material.
    \end{itemize}

\item {\bf Experiment Statistical Significance}
    \item[] Question: Does the paper report error bars suitably and correctly defined or other appropriate information about the statistical significance of the experiments?
    \item[] Answer: \answerYes{} 
    \item[] Justification: In the section about applying the SelectIT on MT LLMs, we do the statistical significance tests in Table 9, which is reported in the appendix.
    \item[] Guidelines:
    \begin{itemize}
        \item The answer NA means that the paper does not include experiments.
        \item The authors should answer "Yes" if the results are accompanied by error bars, confidence intervals, or statistical significance tests, at least for the experiments that support the main claims of the paper.
        \item The factors of variability that the error bars are capturing should be clearly stated (for example, train/test split, initialization, random drawing of some parameter, or overall run with given experimental conditions).
        \item The method for calculating the error bars should be explained (closed form formula, call to a library function, bootstrap, etc.)
        \item The assumptions made should be given (e.g., Normally distributed errors).
        \item It should be clear whether the error bar is the standard deviation or the standard error of the mean.
        \item It is OK to report 1-sigma error bars, but one should state it. The authors should preferably report a 2-sigma error bar than state that they have a 96\% CI, if the hypothesis of Normality of errors is not verified.
        \item For asymmetric distributions, the authors should be careful not to show in tables or figures symmetric error bars that would yield results that are out of range (e.g. negative error rates).
        \item If error bars are reported in tables or plots, The authors should explain in the text how they were calculated and reference the corresponding figures or tables in the text.
    \end{itemize}

\item {\bf Experiments Compute Resources}
    \item[] Question: For each experiment, does the paper provide sufficient information on the computer resources (type of compute workers, memory, time of execution) needed to reproduce the experiments?
    \item[] Answer: \answerYes{} 
    \item[] Justification: In Section 5.3,  we provide the type of computing workers, memory, and time of execution to help other researchers reproduce the Selective Alpaca.
    \item[] Guidelines:
    \begin{itemize}
        \item The answer NA means that the paper does not include experiments.
        \item The paper should indicate the type of compute workers CPU or GPU, internal cluster, or cloud provider, including relevant memory and storage.
        \item The paper should provide the amount of compute required for each of the individual experimental runs as well as estimate the total compute. 
        \item The paper should disclose whether the full research project required more compute than the experiments reported in the paper (e.g., preliminary or failed experiments that didn't make it into the paper). 
    \end{itemize}
    
\item {\bf Code Of Ethics}
    \item[] Question: Does the research conducted in the paper conform, in every respect, with the NeurIPS Code of Ethics \url{https://neurips.cc/public/EthicsGuidelines}?
    \item[] Answer: \answerYes{} 
    \item[] Justification: we have read the guidelines and ensured that our paper conforms to them.
    \item[] Guidelines:
    \begin{itemize}
        \item The answer NA means that the authors have not reviewed the NeurIPS Code of Ethics.
        \item If the authors answer No, they should explain the special circumstances that require a deviation from the Code of Ethics.
        \item The authors should make sure to preserve anonymity (e.g., if there is a special consideration due to laws or regulations in their jurisdiction).
    \end{itemize}

\item {\bf Broader Impacts}
    \item[] Question: Does the paper discuss both potential positive societal impacts and negative societal impacts of the work performed?
    \item[] Answer: \answerYes{}  
    \item[] Justification: In Section 5.1,  Selective can use fewer computing resources to select high-quality data, which has a positive impact on society. We also have a section to discuss the broader impacts.
    \item[] Guidelines:
    \begin{itemize}
        \item The answer NA means that there is no societal impact of the work performed.
        \item If the authors answer NA or No, they should explain why their work has no societal impact or why the paper does not address societal impact.
        \item Examples of negative societal impacts include potential malicious or unintended uses (e.g., disinformation, generating fake profiles, surveillance), fairness considerations (e.g., deployment of technologies that could make decisions that unfairly impact specific groups), privacy considerations, and security considerations.
        \item The conference expects that many papers will be foundational research and not tied to particular applications, let alone deployments. However, if there is a direct path to any negative applications, the authors should point it out. For example, it is legitimate to point out that an improvement in the quality of generative models could be used to generate deepfakes for disinformation. On the other hand, it is not needed to point out that a generic algorithm for optimizing neural networks could enable people to train models that generate Deepfakes faster.
        \item The authors should consider possible harms that could arise when the technology is being used as intended and functioning correctly, harms that could arise when the technology is being used as intended but gives incorrect results, and harms following from (intentional or unintentional) misuse of the technology.
        \item If there are negative societal impacts, the authors could also discuss possible mitigation strategies (e.g., gated release of models, providing defenses in addition to attacks, mechanisms for monitoring misuse, mechanisms to monitor how a system learns from feedback over time, improving the efficiency and accessibility of ML).
    \end{itemize}
    
\item {\bf Safeguards}
    \item[] Question: Does the paper describe safeguards that have been put in place for responsible release of data or models that have a high risk for misuse (e.g., pretrained language models, image generators, or scraped datasets)?
    \item[] Answer: \answerNA{} 
    \item[] Justification: The model and datasets we used are all open-sourced, and we strictly follow their terms once the terms are carried out.
    \item[] Guidelines:
    \begin{itemize}
        \item The answer NA means that the paper poses no such risks.
        \item Released models that have a high risk for misuse or dual-use should be released with necessary safeguards to allow for controlled use of the model, for example by requiring that users adhere to usage guidelines or restrictions to access the model or implementing safety filters. 
        \item Datasets that have been scraped from the Internet could pose safety risks. The authors should describe how they avoided releasing unsafe images.
        \item We recognize that providing effective safeguards is challenging, and many papers do not require this, but we encourage authors to take this into account and make a best faith effort.
    \end{itemize}

\item {\bf Licenses for existing assets}
    \item[] Question: Are the creators or original owners of assets (e.g., code, data, models), used in the paper, properly credited and are the license and terms of use explicitly mentioned and properly respected?
    \item[] Answer: \answerYes{}  
    \item[] Justification: We have cited the creators in the main part of the paper and the supplement material.
    \item[] Guidelines:
    \begin{itemize}
        \item The answer NA means that the paper does not use existing assets.
        \item The authors should citep the original paper that produced the code package or dataset.
        \item The authors should state which version of the asset is used and, if possible, include a URL.
        \item The name of the license (e.g., CC-BY 4.0) should be included for each asset.
        \item For scraped data from a particular source (e.g., website), the copyright and terms of service of that source should be provided.
        \item If assets are released, the license, copyright information, and terms of use in the package should be provided. For popular datasets, \url{paperswithcode.com/datasets} has curated licenses for some datasets. Their licensing guide can help determine the license of a dataset.
        \item For existing datasets that are re-packaged, both the original license and the license of the derived asset (if it has changed) should be provided.
        \item If this information is not available online, the authors are encouraged to reach out to the asset's creators.
    \end{itemize}

\item {\bf New Assets}
    \item[] Question: Are new assets introduced in the paper well documented and is the documentation provided alongside the assets?
    \item[] Answer: \answerYes{}  
    \item[] Justification: In the abstract section, we provide the GitHub link to open access to our code and data.
    \item[] Guidelines:
    \begin{itemize}
        \item The answer NA means that the paper does not release new assets.
        \item Researchers should communicate the details of the dataset/code/model as part of their submissions via structured templates. This includes details about training, license, limitations, etc. 
        \item The paper should discuss whether and how consent was obtained from people whose asset is used.
        \item At submission time, remember to anonymize your assets (if applicable). You can either create an anonymized URL or include an anonymized zip file.
    \end{itemize}

\item {\bf Crowdsourcing and Research with Human Subjects}
    \item[] Question: For crowdsourcing experiments and research with human subjects, does the paper include the full text of instructions given to participants and screenshots, if applicable, as well as details about compensation (if any)? 
    \item[] Answer: \answerNA{} 
    \item[] Justification: SelectIT does not involve crowdsourcing or research with human subjects.
    \item[] Guidelines:
    \begin{itemize}
        \item The answer NA means that the paper does not involve crowdsourcing nor research with human subjects.
        \item Including this information in the supplemental material is fine, but if the main contribution of the paper involves human subjects, then as much detail as possible should be included in the main paper. 
        \item According to the NeurIPS Code of Ethics, workers involved in data collection, curation, or other labor should be paid at least the minimum wage in the country of the data collector. 
    \end{itemize}

\item {\bf Institutional Review Board (IRB) Approvals or Equivalent for Research with Human Subjects}
    \item[] Question: Does the paper describe potential risks incurred by study participants, whether such risks were disclosed to the subjects, and whether Institutional Review Board (IRB) approvals (or an equivalent approval/review based on the requirements of your country or institution) were obtained?
    \item[] Answer: \answerNA{} 
    \item[] Justification: SelectIT does not involve crowdsourcing or research with human subjects.
    \item[] Guidelines: 
    \begin{itemize}
        \item The answer NA means that the paper does not involve crowdsourcing nor research with human subjects.
        \item Depending on the country in which research is conducted, IRB approval (or equivalent) may be required for any human subjects research. If you obtained IRB approval, you should clearly state this in the paper. 
        \item We recognize that the procedures for this may vary significantly between institutions and locations, and we expect authors to adhere to the NeurIPS Code of Ethics and the guidelines for their institution. 
        \item For initial submissions, do not include any information that would break anonymity (if applicable), such as the institution conducting the review.
    \end{itemize}
\end{enumerate}

\end{document}